%% file: main.tex
\documentclass[a4paper,11pt,twocolumn,twoside]{article}
\usepackage{graphicx}
\usepackage{sepln_sty/sepln_en}
\usepackage{sepln_sty/fullname}
\usepackage{multirow}
\usepackage[utf8]{inputenc}
\usepackage{babel} 
\usepackage{amsmath}
\usepackage{fbox}

\title{Contrastive Analysis of Linguistic Representations in Large Language Model Outputs through Structured Synthetic Data Generation and Abstracted N-gram Associations}

\author {\textbf{Sofía A. Desimone,$^{1,2}$} \textbf{Laura Alonso Alemany$^1$}\\
$^1$Universidad Nacional de Córdoba\\
$^2$CONICET
}

\seplntranstitle{Análisis contrastivo de representaciones lingüísticas en salidas de modelos de lenguaje de gran escala mediante generación estructurada de datos sintéticos y asociaciones abstractas de n-gramas}

\seplnclave{Modelos de Lenguaje de Gran Escala,
Análisis de Sesgos,
Datos Sintéticos}

\seplnresumen{Presentamos un marco metodológico para descubrir patrones lingüísticos y discursivos asociados a distintos grupos sociales mediante generación sintética contrastiva de texto y análisis estadístico. A diferencia de enfoques previos, nuestro objetivo es caracterizar formas sutiles de sesgo, en lugar de diagnosticarlo a partir de listas predefinidas de palabras o expresiones. Asimismo, trabajamos con datos contextualizados en lugar de palabras o oraciones aisladas.
Nuestra metodología se aplica a producciones textuales de cualquier género, incluyendo narrativas, textos orientados a tareas o interacciones dialógicas. Los datos contextualizados se generan mediante combinaciones controladas de escenarios situacionales y marcadores de grupo, creando pares mínimos de textos que difieren únicamente en el grupo referido, mientras mantienen condiciones comparables.
Para facilitar un análisis robusto, las formas lingüísticas se generalizan y las asociaciones entre abstracciones lingüísticas y grupos se cuantifican mediante una variante de la información mutua puntual, lo que permite detectar expresiones que aparecen de manera desproporcionada entre grupos. Finalmente, una estrategia de priorización de fragmentos identifica segmentos textuales con alta concentración de señales lingüísticas potencialmente sesgadas, lo que permite a expertos evaluar el posible daño de dichas expresiones en contexto, articulando el análisis cuantitativo con la interpretación cualitativa.}

\seplnkey{Large Language Models,
Bias Analysis,
Synthetic Data}

\seplnabstract{We present a methodological framework to discover linguistic and discursive patterns associated to different social groups through contrastive synthetic text generation and statistical analysis. In contrast with previous approaches, we aim to characterize subtle expressions of bias, instead of diagnosing bias through a pre-determined list of words or expressions. We are also working with contextualized data instead of isolated words or sentences. Our methodology applies to textual productions in any genre, encompassing narrative, task-oriented or dialogic.   
Contextualized data are generated using controlled combinations of situational scenarios and group markers, creating minimal pairs of texts that differ only in the referenced group while maintaining comparable narrative conditions.
To facilitate robust analysis, linguistic forms are generalized 
and associations between linguistic abstractions and groups are quantified using a 
variant of pointwise mutual information to detect expressions that appear disproportionately across groups. A fragment-ranking strategy then prioritizes text segments with a high concentration of biased linguistic signals, which allows for experts to assess the harmful potential of linguistic expressions in  context, bridging quantitative analysis and qualitative interpretation.
}

\firstpageno{1}

\begin{document}

\maketitle

\section{Introduction and Motivation}

\input{secciones/intro}

\section{Relevant Work}

\input{secciones/relevant_work}

\section{Contrastive Text Generation}

\input{secciones/story_generation}

\section{Discovery of Associations}

\input{secciones/generalization}

\subsection{Measuring strength of association}

\input{secciones/association}

\section{Fragment Ranking}

\input{secciones/ranking}

\section{Proof of Concept with Disability in Spanish}

\input{secciones/proof_of_concept_disability}

\section{Discussion}

\input{secciones/discussion}

\bibliographystyle{sepln_sty/fullname}
\bibliography{subtle_biases.bib}

\end{document}

%% file: secciones/intro.tex
Large Language Models (LLMs) have demonstrated outstanding performance in text generation tasks. However, numerous studies have pointed out that these systems may reproduce and amplify biases present in their training data, particularly along dimensions such as gender, race, sexual orientation, and religion  \cite{caliskan2017,mehrabi2021,tamkin2023,gallegos2024,caliskan2017,mehrabi2021,tamkin2023,kotek2023,gallegos2024}. In a context where LLMs are rapidly being integrated into high-impact social domains, such as healthcare, education, and justice, the rigorous evaluation of their discursive behavior becomes a priority issue \cite{bender2021,wan2023,panda2025}.

Much of the literature has focused on explicit forms of bias, in which marginalized groups are directly named and associated with identifiable stereotypes. However, bias may also operate implicitly, through more subtle and indirect mechanisms that shape model outputs\cite{kumar2024,bai2025,hirsch2026}. These forms of bias often manifest through complex textual patterns rather than overtly negative statements, including differences in the attribution of roles, agency, or evaluative traits. For instance, certain groups may be disproportionately framed through narratives that are superficially positive yet ultimately limiting, such as overly inspirational or patronizing portrayals, while others are more frequently associated with attributes like professionalism, competence, or precision \cite{grue2016,viveros2016}. Such patterns may also emerge through differential associations with activities, emotional registers, or social roles, including both overrepresentation and systematic absence

These manifestations are inherently more difficult to detect, particularly when analysis relies on predefined lists of sensitive terms, toxicity metrics, or fixed categorical frameworks (e.g., emotion labels such as fear, joy, or positivity). While useful in specific settings, these approaches tend to fragment the phenomenon and impose a reductionist structure that may obscure the broader discursive configurations through which subtle biases are expressed\cite{miceli2025,panda2025}.

In this work, we propose a methodology aimed at identifying subtle, systematically differentiated patterns in generated text through contrastive synthetic data and statistical analysis of associations. Our approach seeks to identify textual fragments that concentrate linguistic configurations differentially associated with social groups under equivalent contextual conditions. These fragments are then assessed by experts to determine whether they may produce harms. The workflow consists of three main stages: (1) contrastive generation of texts for each social group; (2) statistical discovery of association patterns between social groups and textual expressions; and (3) identification of textual fragments with a high density of differentially associated expressions, so that experts may assess whether they produce any kind of harm. This design allows experts to focus on fragments that exhibit strong differential behaviors, facilitating the identification and characterization of discursive patterns associated with unequal or potentially harmful representations. In contrast to previous approaches, which often require inspecting large volumes of text or decontextualized outputs, our method directs attention to a reduced set of context-rich and statistically grounded examples.

The rest of the paper is organized as follows. In the next Section, we describe the capabilities and limitations of different approaches to bias detection, and how this approach proposes to overcome them. Then, we describe the main parts of the proposed methodology: contrastive generation of texts, discovery of associations and fragment ranking and assessment. Finally, we present a proof of concept with the group of people with disabilities in Spanish.

%% file: secciones/relevant_work.tex
Research on bias in language models has received increasing attention due to both the technical challenges it poses and its social implications \cite{bender2021,gallegos2024,kumar2024}. Numerous studies have developed methods to explore, quantify, and mitigate bias in automated linguistic systems, consolidating an active field of evaluation and auditing.

A substantial portion of this literature has focused on detecting explicit bias, frequently through the use of predefined lists of sensitive terms, stereotypical templates, or structured benchmarks \cite{zhao2018,nangia2020,nadeem2021,parrish2022}. While these approaches have proven effective in identifying overt forms of discrimination, they present limitations in capturing more nuanced phenomena, such as differential narrative framing, implicit condescension, or systematically biased representations that are not expressed through openly offensive language.

In this regard, recent work emphasizes that some of the most persistent forms of bias in LLMs are implicit or structural \cite{kumar2024,panda2025,bai2025,hirsch2026,bali-etal-2026-detecting}, operating through recurrent associations and discursive patterns that manifest indirectly. In response to these limitations, more contextualized approaches have begun to emerge, 
aimed at identifying patterns that arise beyond predefined categories.

In this work, we propose a methodology aimed at operationalizing qualitative evaluation and collective participation, systematically producing evidence that can be critically examined by domain experts and potentially affected communities. In this way, the methodology structures the exploration process and the detection of statistically significant patterns, while experts are only required to assess the potential harms present in the identified fragments.

In contrast with other approaches, our method does not predefine bias indicators; instead, these are discovered from the texts themselves and are not constrained to any fixed or predefined form. Rather, they may take arbitrary linguistic configurations. Moreover, expert involvement always occurs at a level of abstraction or concreteness that is natural to them and in a contextualized manner: by describing groups and situations that may be differentially represented by LLMs, and by evaluating the potential harm in fragments embedded within complete texts.

%% file: secciones/story_generation.tex
The first step in our methodology consists in generating synthetic data to obtain controlled contexts in which differential behavior across social groups may emerge. By constructing minimal pairs of prompts that vary only in the social group referenced, the methodology enables the systematic analysis of whether and how LLMs produce textual outputs for different groups under equivalent conditions. 

This approach is based on the assumption that subtle biases often emerge only through comparison: linguistic patterns that appear neutral in isolation, like the word "\textit{inspiring}", may reveal  discriminatory effects when we can systematize how they are differentially distributed across texts for different social groups, and how they may occupy the space of other expressions and concepts, like  "\textit{reliable}". 

The process is organized around two main components: (i) a set of scenarios that define the situational context and remain constant across conditions, usually originating from  initial hypotheses  from experts; and (ii) a set of markers that specify attributes of a person, distinguishing between the group of interest and the control groups.

From a base prompt like the one shown in Figure \ref{fig:example_prompt}, pairs of texts are generated that share the same situational context but differ only in the marker used. Thus, we obtain minimal pairs of texts. The methodology of minimal pairs has been extensively used in bias measurements (e.g.,  Nangia et al. \shortcite{nangia2020}), but these minimal pairs are crucially contextualized, in contrast with isolated words or sentences. This contrastive design enables direct comparison between comparable narratives because the texts resulting from minimal pairs of prompts are comparable texts, facilitating the identification of patterns that occur disproportionately in texts associated with the group of interest. This approach makes it possible to detect subtle biases that emerge at the level of narrative organization and may remain invisible under purely lexical analysis. At the same time, it allows experts to describe their hypotheses on discrimination in a natural way.

%% file: secciones/generalization.tex
Once the texts have been generated through a contrastive process, we apply statistical analysis to determine whether they differ significantly across the groups under comparison, and to characterize these differences when they arise. To this end, we aim to identify textual expressions that are more strongly associated with one group than with others. When such expressions exhibit a stronger association with a particular group, we interpret them as evidence of differential model behavior and consider them as indicators of group-specific patterns.

A simple and intuitive measure of association is Pointwise Mutual Information (PMI), which has been successfully linked to various bias detection approaches \cite{bordia-bowman-2019-identifying,aka-etal-21-bias-no-ground-truth,valentini-etal-2023-interpretability}. However intuitive, the main shortcoming of PMI is that it is sensitive to the frequency of the analyzed units, which makes it difficult to compare associations when dealing with low-frequency expressions. In particular, PMI estimates tend to be less stable for rare events, which may affect the interpretability of the results. For this reason, in order to make the measure less sensitive to frequency, we propose to generalize linguistic expressions by grouping low-frequency expressions into more general units that share a common meaning. Once this generalization is performed, PMI can be applied to these units. In the remainder of this section, we describe how linguistic forms are generalized and how associations between these generalized forms and the groups under study are identified.

\subsection{Generalization of Linguistic Forms}

To mitigate the sparsity of the distribution and reduce the impact of low-frequency events, we generalize linguistic forms by constructing more abstract analytical units. First, all tokens except stopwords are lemmatized, which reduces morphological variability and decreases the number of distinct units under analysis. 

Building on this representation, we define equivalence classes of n-grams. For each n-gram, non-stopword tokens are mapped to their corresponding lemmas. N-grams that share the same set of lemmas are then grouped into a common equivalence class, independently of word order. This grouping relies on the assumption that such n-grams convey a similar underlying meaning despite surface-level variation. By aggregating multiple low-frequency realizations into shared abstractions, this process produces a more stable distribution over which association measures such as PMI can be more reliably estimated.

An example can be seen in Figure \ref{fig:example_equivalence_classes}, where all n-grams belong to the same equivalence class because their content words correspond to the lemmas “enfrentar” (\textit{face}) and “obstáculo” (\textit{obstacle}).

\begin{figure}
    \centering
    \fbox{
    \begin{minipage}{.45\textwidth}
    “enfrentado todos los obstáculos”, “obstáculos que había enfrentado”, “enfrentado varios obstáculos”, “enfrentó varios obstáculos”, “enfrentas obstáculos”, “enfrentado obstáculos”.
    \end{minipage}
    }
    \caption{Equivalence class of n-grams with the lemmas “enfrentar” (\textit{face}) and “obstáculo” (\textit{obstacle}).}
    \label{fig:example_equivalence_classes}
\end{figure}

We exclude n-grams bebinning or ending with a stopword. For example, the trigram “amigos y familia” (\textit{friends and family}) would be included, whereas “los amigos y” (\textit{the friends and}) would be excluded. This is because n-grams that begin or end with a stopword tend to lack semantic autonomy and provide less stable analytical units.


%% file: secciones/association.tex
Once linguistic forms have been generalized into equivalence classes, we measure the strength of association between each equivalence class and each of the groups under study. The underlying intuition is that if an equivalence class occurs more frequently in texts generated for one of the groups than in texts generated for any other group, then it can be considered more strongly associated with that group, with a strength proportional to the difference in the number of occurrences across groups. To do that, we compute the PMI between equivalence classes of n-grams and the groups.

Formally, the PMI between a linguistic unit $w$ and a group $G$ is defined as:
\begin{equation}
\text{PMI}(w, G) = \log \frac{P(w \mid G)}{P(w)}.
\end{equation}

With this definition of PMI, and comparably to the intuition captured by previous work like Valentini \shortcite{valentini-etal-2023-interpretability}, we can define the \textit{BiasScore} (BS) of a unit $w$ as the difference between its PMI with respect to a group of interest ($GI$) and a control group ($GC$):
\begin{align}
\text{BS}(w)
&= \text{PMI}(w, GI) - \text{PMI}(w, GC) \\
&= \log \frac{P(w \mid GI)}{P(w \mid GC)}.
\end{align}

In practice, BS is estimated from frequency counts: 
\begin{equation}
\text{BS}(w) = \log \frac{M_{w,GI}}{M_{\cdot GI}} - \log \frac{M_{w,GC}}{M_{\cdot GC}}
\end{equation}
where $M_{w,G}$ denotes the frequency of $w$ in group $G$, and $M_{\cdot G}$ the total number of units observed in that group. 

Our analysis operates over equivalence classes of n-grams. The frequency of an equivalence class is defined as the sum of the frequencies of all units that belong to it.

For higher-order n-grams, we treat the ratio 
\(\frac{M_{\cdot GI}}{M_{\cdot GC}}\) 
as constant across different equivalence classes, as it primarily depends on the relative size of the corpora in each group. This allows us to apply the same normalization scheme across different types of units.

An example of the BS assigned to different equivalence classes can be seen in Tables \ref{tab:equiv_highBias} and \ref{tab:equiv_lowbias}.

%% file: secciones/ranking.tex
Finally, once we have discovered the strength of association of textual expressions with each group, we can use this information to identify fragments of text that contain a higher concentration of expressions more strongly associated with a given group relative to others, such as those depicted in Figure \ref{fig:fragments}. These fragments are then assessed by experts to determine whether they cause harm.

In this stage, we operationalize the identification of potentially biased text fragments by leveraging the previously defined equivalence classes and their corresponding $BS$ values. This involves specifying which types of fragments and equivalence classes are relevant, as well as applying a scoring mechanism to prioritize fragments according to the concentration of differential associations they contain. While the details of this process can be adapted to the goals of each analysis, here we propose a strategy that serves as a general methodological framework, flexible across different corpora, application contexts, and research questions.

\subsection{Filtering intrinsic differences}

Some expressions may show a strong association with a given group because they refer to intrinsic characteristics of that group, and not necessarily to indicators of bias. In these cases, the differential distribution of certain n-grams across groups does not constitute evidence of subtle bias, but rather reflects inherent or expected properties.

Consequently, not all equivalence classes derived from the generalization process are equally informative for analysis. To address this issue, we construct a set of expressions considered inherent to the groups under study and exclude the equivalence classes that contain them. For example, if the group of interest is people with disabilities, the word "disléxico" (\textit{dyslexic}) would lead to the exclusion of equivalence classes containing n-grams with forms such as "disléxica" (feminine), "disléxicos", or "disléxicas".

Additionally, filters can be applied based on lists defined by domain experts, individuals with lived experience, or other relevant stakeholders. These lists include expressions whose differential occurrence across groups is expected and does not, by itself, indicate bias (e.g., “cane” or “sign language”). This step allows the analysis to focus on subtle linguistic patterns rather than on terminology explicitly linked to identity.

Furthermore, in order to mitigate the effects of low frequency and ensure the relevance of the analyzed units, we introduce minimum frequency thresholds and restrict the analysis to equivalence classes whose frequency exceeds a given value. Units with very low occurrences tend to produce unstable estimates and lack statistical relevance. Since units with a higher number of content-bearing tokens are naturally less frequent, thresholds are set differently according to the number of non-stopword tokens in each unit.

\subsection{Definition of Candidate Fragments}

We define a set of fragments that meet previously established conditions. These conditions aim to delimit comparable textual units that are relevant for analysis. In particular, they may include structural constraints such as a fixed number of sentences, the presence of explicit markers associated with the group of interest, or other relevant criteria.

Based on this set of fragments, a scoring scheme is constructed to estimate the concentration of linguistic patterns that are differentially associated with the groups.

\subsection{Fragment Scoring}

In order to identify fragments that concentrate a higher density of bias, we propose a scoring scheme based on the $BS$ values calculated over the previously defined equivalence classes. This score allows us to establish a priority ranking among fragments according to their degree of differential association with the groups under comparison.

Formally, given a fragment $F$, we consider the n-grams appearing in it and associate them with their respective equivalence classes. Under the assumption that longer units provide greater contextual specificity, we avoid double counting by discarding n-grams that are completely contained within longer n-grams present in the same fragment and at the same position.

The fragment score $Score(F)$ is defined as the sum of the 
$BS$ values of the equivalence classes associated with the n-grams in the fragment that belong to equivalence classes considered in the analysis. Formally, let 
$a$ denote an equivalence class associated with an n-gram in 
$F$ then:
\[
Score(F) = \sum BS(a)
\]
where the sum runs over the occurrences of n-grams in $F$ that are associated with equivalence classes that were not filtered. In particular, when multiple n-grams belonging to the same equivalence class appear in different positions of the fragment, their contributions are accumulated. In cases where $BS(a)$ is not finite, its value is replaced by the finite value with the largest absolute magnitude observed among the equivalence classes considered in the analysis (i.e., those that were not filtered).

It is important to note that frequency thresholds are adjusted according to the number of content-bearing tokens in each equivalence class. Classes with more words tend to be less frequent and, therefore, exhibit greater variability in BS; however, these longer units provide more specific and relevant information to identify differential discursive patterns.

Consequently, fragments are ranked according to the density of linguistic units differentially associated with the groups under comparison. Fragments ranked highest are more strongly associated with the group of interest, whereas fragments ranked lowest are more strongly associated with the control group. It is important to emphasize that this ranking mechanism allows expert evaluation to focus on segments most likely to contain relevant linguistic patterns.

%% file: secciones/proof_of_concept_disability.tex
As a proof of concept, we have carried out an initial exploration on representations related to persons with disabilities. Unlike other bias dimensions that have been more extensively studied, disability has received comparatively less attention in LLM analysis, and associated biases often manifest in particularly subtle and narratively mediated ways \cite{phutane2025,panda2025}. This case illustrates the potential of the method to detect discursive patterns that do not emerge through isolated lexical analyses.

To explore the behavior of the proposed methodology, we generated a set of stories in Spanish using the model \texttt{llama-3.3-70B}. The corpus was organized into contrastive pairs: in one set of stories the character is explicitly described as having a disability, while in the control set the same narrative context is maintained without that reference. In both cases, the scenario and narrative task remain constant, and only the marker associated with the character is modified. To do that, we used prompts such as those in Figure \ref{fig:example_prompt}.

\begin{figure*}
    \centering
    \fbox{
    \begin{minipage}{\linewidth}
        Escribe una historia sobre Ana, \textbf{una persona ciega} que tiene una reunión de trabajo.\\
        \textit{Write a story about Ana, \textbf{a blind person} who has a work meeting.}
        
        Escribe una historia sobre Ana, \textbf{una persona con anteojos} que tiene una reunión de trabajo.\\
        \textit{Write a story about Ana, \textbf{a person wearing glasses} who has a work meeting.}
    \end{minipage}}
    \caption{An example prompts to generate contrastive stories with constant scenarios (in this case, a work meeting) with a person with disability (in this case, a blind person) and a person with no disability (in this case, a person wearing glasses), marked in boldface in the example.}
    \label{fig:example_prompt}
\end{figure*}

For fragment prioritization, we first defined candidate fragments consisting of three sentences extracted from the stories associated with the group of interest. We additionally required that the central sentence contain a term belonging to the disability group, in order to analyze the immediate discursive context in which this marker is introduced.

Regarding the minimum frequency thresholds for including equivalence classes in the analysis, we set 40 occurrences for unigrams, 20 for classes with two non-stopword tokens, 15 for those with three non-stopword tokens, and 10 for classes with four non-stopword tokens. These values reflect that longer units naturally tend to occur less frequently. To reduce statistical noise, we discarded equivalence classes whose absolute BS value was lower than $0.5$, a threshold corresponding approximately the standard deviation observed in the distribution of values obtained from random corpus partitions.

To illustrate the type of discursive patterns identified by the proposed methodology, we present two example fragments in Figure \ref{fig:fragments}. The first corresponds to a fragment associated with a high positive BS value, while the second corresponds to a fragment with a negative BS value.

\begin{figure*}
\begin{minipage}[t]{\columnwidth}
\textbf{Fragment with high positive BS}

\vspace{0.3em}

\textit{Ana demostró que, con determinación y el adecuado apoyo, cualquier persona puede superar los obstáculos y alcanzar sus metas, independientemente de las barreras que pueda encontrar. Su historia se convirtió en una inspiración para muchos, mostrando que la discapacidad no es un límite para el éxito, sino una parte de la diversidad humana que enriquece nuestras comunidades. Después de graduarse, Ana continuó su camino hacia el éxito, trabajando en proyectos innovadores y contribuyendo a hacer del mundo un lugar más inclusivo y accesible para todos.}
\end{minipage}
\hspace{0.5cm}
\begin{minipage}[t]{\columnwidth}
\textbf{Fragment with high negative BS}

\vspace{0.3em}

\textit{Se detuvo un momento para disfrutar del paisaje y luego continuó su camino por el sendero principal. La silla de ruedas se deslizaba suavemente sobre la superficie plana, permitiéndole avanzar con facilidad. A medida que avanzaba, Juan se encontró con otros visitantes del parque: familias con niños que jugaban en los espacios de juego, parejas que paseaban de la mano y amigos que compartían risas y conversaciones.}
\end{minipage}
\caption{\label{fig:fragments}Example of fragments of texts with high positive BiasScore (strongly associated to the interest group, in this case, people with disabilities) and high negative BiasScore (strongly  associated with the control group, in this case, people without any explicit mention of disabilities).}
\end{figure*}

We observe that the fragment with the highest Score presents a type of representation that may be problematic, particularly in the form of \textit{inspirational porn} \cite{gadiraju2023,li2024}. The equivalence classes contributing most to the Score include expressions emphasizing personal overcoming and inspiration ("cualquier persona puede superar" \textit{anyone can overcome}, "convirtió en una inspiración" \textit{turned into an inspiration}, "alcanzar sus metas" \textit{reach their goals}), suggesting the presence of this type of discriminatory representation. In contrast, the fragment with a negative Score contains expressions centered on everyday experiences ("compartían risas" \textit{shared laughs}, "disfrutar del paisaje" \textit{enjoy the scenery}), which do not emphasize individual effort or personal abilities.


\begin{table*}[t]
\centering
\small

{
\renewcommand{\arraystretch}{1.15}
\setlength{\tabcolsep}{6pt}

\begin{tabular}{p{0.78\textwidth} r}
\hline
\textbf{Equivalence Class} & \textbf{BS} \\
\hline

"adecuado y la determinación", "adecuada y la determinación", "determinación y el adecuado", "adecuadas y la determinación", "determinación adecuada" & 4.42 \\

"barreras que podían", "barreras que podría", "barreras que puedan", "barrera que pudiera", "barreras que pueda", "barreras que podrían", "barrera puede", "barrera podía", "barreras podrían", "barreras podían", "barreras pueden" & 4.42 \\

"inclusivo y accesible", "accesible y inclusivo", "inclusiva y accesible", "accesible e inclusiva", "accesible y inclusiva" & 4.42 \\

"adecuada y el apoyo", "adecuadas y el apoyo", "adecuado y el apoyo", "apoyo adecuado", "adecuado apoyo", "apoyo adecuados" & 4.42 \\

"cualquier persona puede superar" & 3.38 \\

"convirtió en una inspiración", "convertiría en una inspiración", "convirtió en inspiración" & 1.57 \\

"obstáculos y alcanzar", "obstáculo y alcanzar", "obstáculo para alcanzar", "obstáculos y alcanzando", "obstáculos para alcanzar" & 1.13 \\

"encontrar que podía", "pueden encontrar", "podía encontrar", "podría encontrar", "pudiera encontrar", "poder encontrar", "pudo encontrar", "podido encontrar", "podemos encontrar", "puede encontrar", "pueda encontrar", "puedo encontrar", "podrían encontrar", "podían encontrar", "puedan encontrar" & 1.05 \\

"convirtió en una historia", "historia se convirtió", "historia se convertiría" & 1.03 \\

"muchas metas que alcanzaría", "alcanzar sus metas", "alcanzar metas tan", "metas que alcanzaría", "alcanzar su meta", "alcanzó su meta", "alcanzó las metas", "alcanzar metas", "alcanzando metas", "alcanzaran esas metas" & 1.00 \\

"sino como una parte", "sino también una parte", "sino una parte", "sino como parte", "sino parte" & 0.96 \\

"lugar en el mundo", "mundo un lugar", "lugares y mundos" & 0.89 \\

"hacía que el mundo", "hace que el mundo", "hacer que el mundo", "mundo la hizo", "hacer del mundo", "haciendo del mundo", "mundo y hacer", "hacer el mundo", "mundo lo hizo", "mundo haciendo" & 0.74 \\

\hline
\end{tabular}
}

\caption{\label{tab:equiv_highBias}Equivalence classes contributing to the high positive BS fragment in Figure \ref{fig:fragments} (expressions more strongly associated to the group of people with disabilities).}
\end{table*}

\begin{table*}[t]
\centering
\small

{
\renewcommand{\arraystretch}{1.15}
\setlength{\tabcolsep}{6pt}

\begin{tabular}{p{0.78\textwidth} r}
\hline
\textbf{Equivalence Class} & \textbf{BS} \\
\hline

"compartir tanto las risas", "compartir una risa", "risas que compartiría", "compartieron una risa", "compartir risas", "compartían risas", "compartiendo risas", "risas compartieron", "compartieron risas", "compartió risas" & -0.97 \\

"conversación y la risa", "risas y las conversaciones", "risa y las conversaciones", "risas y conversaciones", "risas y conversación"& -0.65 \\

"disfrutar de ese momento", "disfrutar de su momento", "disfrutaron de ese momento", "disfrutar de un momento", "disfrutando del momento", "disfrutaba del momento", "momento de disfrutar", "momento para disfrutar", "disfrutar los momentos", "disfrutar del momento", "disfrutar de momentos", "disfrutó del momento", "disfrutaron del momento", "disfrutar momentos" & -0.57 \\

"disfrutar del paisaje", "disfrutando del paisaje", "disfrutaba del paisaje", "disfrutó del paisaje" & -0.57 \\

\hline
\end{tabular}
}

\caption{\label{tab:equiv_lowbias}Equivalence classes contributing to the negative BS fragment in Figure \ref{fig:fragments} (expressions more strongly associated to the group of people without any explicit mention of disabilities).}
\end{table*}

As shown in Tables \ref{tab:equiv_highBias} and \ref{tab:equiv_lowbias}, the equivalence classes associated with each group help identify specific linguistic associations that differentiate these narratives. These classes serve as valuable indicators, both independently and as contributors to the detection of text fragments that may contain biased or potentially harmful content.

It is important to note that some equivalence classes may group n-grams that, although sharing the same lemmas, combine them in ways that produce different meanings. For example, the class that includes expressions such as "adecuado y la determinación" (\textit{adequate and the determination}), "determinación adecuada" (\textit{adequate  determination}), and "determinación y el adecuado2 (\textit{determination and the adequate}) combines the same lemmas ("adecuado" and "determinación"), but the word order slightly changes the focus or intent of the expression. Even so, we consider that using equivalence classes is preferable to analyzing individual n-grams, as it allows n-grams sharing a common underlying idea to be grouped into a single unit of analysis, reducing dispersion and facilitating the detection of linguistic patterns.

%% file: secciones/discussion.tex
In this work, we present a methodology for the analysis of discursive bias in language models, combining contrastive generation, statistical analysis, and the retrieval of contextualized fragments. In particular, the approach shifts the focus from isolated lexical units to broader discursive configurations, facilitating the identification of patterns that emerge at the narrative level.

A central aspect of the proposed approach is the use of equivalence classes as analytical units. These classes reduce morphological variability and organize information in a way that supports both quantitative analysis and qualitative interpretation. They also help make explicit the linguistic signals underlying the detected patterns, facilitating the interpretation of results and their subsequent evaluation by domain experts.

It is important to note that the construction of these equivalence classes presents certain limitations. In their current form, some classes may be fragmented or group expressions whose semantic relationship is not fully homogeneous, which may affect both the precision of the analysis and its interpretability. Nevertheless, these units enable the identification of regularities that are unlikely to emerge when considering n-grams in isolation. In this regard, future work could explore strategies to improve grouping criteria and strengthen the internal consistency of these classes.

Overall, these elements open up avenues for future research aimed at refining analytical units, evaluating the robustness of the approach across domains, and advancing validation frameworks that systematically incorporate expert participation.

This methodology is specifically aimed to operationalize qualitative evaluation and collective participation, systematically producing evidence that can be critically examined by domain experts and potentially affected communities. Thus it is not experts that need to systematize, but they just need to exercise their judgement. It is the methodology that systematizes the exploration, the detection of statistically significant patterns, and experts only need to determine the possible harms in story fragments.

In contrast with other approaches, this approach does not pre-determine which are indicators of bias, but they are discovered from texts, they do not have a pre-determined form but can be arbitrary in form, and the intervention of experts is always at a level of abstraction or concretion that feels natural to them, and  contextualized: in describing groups and situations that may be portrayed differently by llms for different groups, producing potential harms, and in judging the harm produced by fragments that are contextualized within full texts. In contrast with fully contextualized harms assessments, that usually require immersion in a given context or task, this methodology is less costly, more generalizable, and allows for prospective analysis before deployment.